\documentclass[conference]{IEEEtran}
\IEEEoverridecommandlockouts
\usepackage{cite}
\usepackage{amsmath,amssymb,amsfonts}
\usepackage{algorithmic}
\usepackage{graphicx}
\usepackage{textcomp}
\usepackage{xcolor}
\usepackage{times}
\usepackage{hyperref}

\def\BibTeX{{\rm B\kern-.05em{\sc i\kern-.025em b}\kern-.08em
    T\kern-.1667em\lower.7ex\hbox{E}\kern-.125emX}}

\makeatletter
\renewcommand\IEEEkeywordsname{Keywords}
\makeatother
\usepackage{xpatch}
\xpatchcmd\IEEEkeywords{---}{-}{}{}

\makeatletter
\renewcommand{\fnum@figure}{Figure~\thefigure}
\makeatother

\begin{document}

\title{\bfseries\Large Sequence Graph Network for Online Debate Analysis}

\author{
    \IEEEauthorblockN{
        Quan Mai, \textsuperscript{1}
        Susan Gauch, \textsuperscript{1}
        Douglas Adams, \textsuperscript{2}
        Miaoqing Huang \textsuperscript{1}
        }
    \IEEEauthorblockA{
        \textit{
                \textsuperscript{\rm 1}Department of Electrical Engineering and Computer Science,
                \textsuperscript{\rm 2}Department of Sociology and Criminology
        } \\
    \textit{University of Arkansas}\\
    Fayetteville, Arkansas, USA \\
    \{quanmai, sgauch, djadams, mqhuang\}@uark.edu}
}

\maketitle

\begin{abstract}
Online debates involve a dynamic exchange of ideas over time, where participants need to actively consider their opponents' arguments, respond with counterarguments, reinforce their own points, and introduce more compelling arguments as the discussion unfolds. Modeling such a complex process is not a simple task, as it necessitates the incorporation of both sequential characteristics and the capability to capture interactions effectively. To address this challenge, we employ a sequence-graph approach. Building the conversation as a graph allows us to effectively model interactions between participants through directed edges. Simultaneously, the propagation of information along these edges in a sequential manner enables us to capture a more comprehensive representation of context. We also introduce a Sequence Graph Attention layer to illustrate the proposed information update scheme. The experimental results show that sequence graph networks achieve superior results to existing methods in online debates.
\end{abstract}


\section{Introduction}
Online debate has become an integral part of our digital age, transforming the way we engage in discourse and exchange ideas. In social media platforms (e.g., Facebook, Twitter (currently X), etc.), individuals from diverse backgrounds and geographical locations converge to discuss and deliberate on a wide array of topics, ranging from politics and ethics to music and science. Debating with a wide range of debaters requires participants to research and present well-informed arguments, encourages critical thinking, and challenges preconceived notions. 

Like other forms of debate, online discussions are contingent on the flow of time (temporal dependency); each subsequent comment relies on the content of the previous comment it responds to.  Participants interactively promote their point while countering the opponent's \cite{zhang2016conversational}.  Within a turn, debaters employ a variety of strategies, each of which plays a crucial role in determining the outcome of the debate. These strategies involve either directly addressing the opponent's argument, presenting their own viewpoint, or skillfully combining both tactics. The latter approach often appears to be the most effective, allowing the debater to simultaneously achieve both objectives during their turn. However, one cannot always adopt that strategy as it depends on their position in the debate. For instance, if a debater is the first speaker in a debate, their primary task is to present their own ideas coherently and logically, as they do not have the opportunity to directly counter their opponent's arguments at this stage. In such a scenario, the debater's effectiveness lies in the clarity and persuasiveness of their presentation, making it challenging for the opposing side to refute their position. These strategies are also discussed in ~\cite{zhang2016conversational}, which examined the dynamics of information flow within online debates.

\begin{figure}
    \centerline{\includegraphics[scale=0.7]{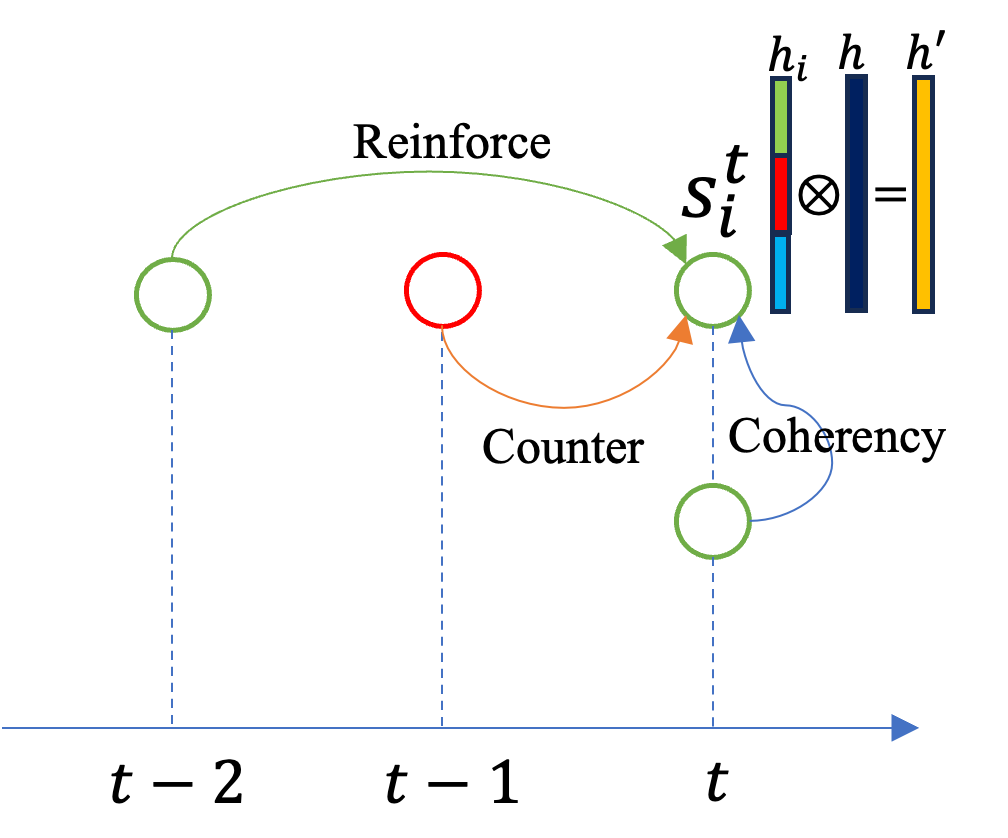}}
    \caption{ A “what-should-we-mention" information flow scheme that mimics the interaction process of a debater. At each time step $t$, the node features are updated by considering their peer nodes from the same turn and the connected nodes from previous turns, using Directed Graph Attention Network layers. Nodes associated with different debaters are colored differently. Each type of edge (colored arrows) contributes a corresponding representation, collectively forming $\mathbf{h}_i$. The node's utterance embedding $\mathbf{h}$ and the interaction representation $\mathbf{h}_i$ are used to update the node feature $\mathbf{h}'$.}
    \label{fig:flow}
\end{figure}

As the argument process is temporally dependent, Recurrent Neural Networks (RNNs), such as Long Short Term Memory (LSTM) \cite{lstm} and Gated Recurrent Unit (GRU) \cite{cho2014learning}, have been one of the most widely used techniques in argument-winning research as well as dialog extraction. Several studies employ RNNs as the encoder for utterances \cite{bao2021argument}\cite{jo2018attentive}\cite{ji2018incorporating}, leveraging their capacity to capture sequential dependencies and relationships within textual data. In addition to encoding individual utterances, sequence networks are employed to encode entire conversations by sequentially processing the arguments \cite{hidey2018persuasive}. 

In a debate, however, participants engage in interactive turn-by-turn rebuttals to counter their opponents' arguments, and sequencing the entire conversation fails to capture this dynamic interaction. In order to model the process of dialogical argumentation,  \cite{ji2018incorporating} use a co-attention network to capture the interaction between the participants and achieve a promising performance on the prediction task. The focus of \cite{jo2018attentive} is placed on identifying connections between the sentences of debaters. This approach is instrumental in capturing critical argumentative components, making it a pivotal factor for predicting the winner. The aforementioned studies compute ``attention scores" for each pair of sentences belonging to two participants in order to assess the \textit{relevance} of one sentence to another. 

An alternative method for capturing these interaction dynamics is through the use of graphs.  Graphs are an effective way to represent relationships and dependencies among entities, making them suitable for a wide range of applications, including social networks and recommendation systems \cite{wu2022graph, fan2019graph, cao2020popularity}. The connection between two components of an argument can be effectively represented by a link (or edge) within the graph. Graphs can also serve as input to Graph Neural Networks (GNNs) for capturing the contextual information within the conversation. In their work, \cite{chen2023dialogue} employs a heterogeneous graph to represent the relationships among entities discussed in multi-party dialogues. In order to model the relationships between argument pairs, \cite{bao2021argument} incorporate intra-passage and cross-passage links to interconnect sentence nodes. Subsequently, they employ a Graph Convolutional Network (GCN) \cite{kipf2017semi} for efficient information propagation. 

Traditional GNNs, including GCNs and Graph Attention Networks (GAT) \cite{velivckovic2017graph}), may not effectively capture the temporal dynamics within a conversation, particularly in a debate scenario in which participants engage in interactive exchanges to counter arguments or defend their own viewpoints. To tackle this challenge, we integrate the strengths of both RNNs and GNNs within a unified framework. In this framework, we conceptualize the debate as a graph, where argument components are depicted as nodes, and their features undergo sequential updates, according to the turn to which they correspond. We introduce the Sequence Graph Attention (SGA) cell, which resembles the traditional RNN-cell, to capture long-range dependencies in the debate (which is treated as a sequence of subgraphs). The experimental results demonstrate that our approach can capture the interaction between debaters and outperforms state-of-the-art models in accurately predicting the winner in several online debate datasets. The code and models are available at \url{https://github.com/quanmai/SGA}

The structure of the remainder of this paper is organized as follows: Section \ref{sec: preliminary} describes the process of constructing a graph from a debate. In Section \ref{sec: method}, we introduce our proposed framework. The effectiveness of this method is evaluated in Section \ref{sec:evaluation}. Section \ref{sec: related} reviews some relevant literature. Finally, Section \ref{sec: conclusion} provides a summary of our findings and discusses potential avenues for future work.

\section{Preliminary} \label{sec: preliminary}
\begin{figure*}
    \centering
    \includegraphics[scale=0.6]{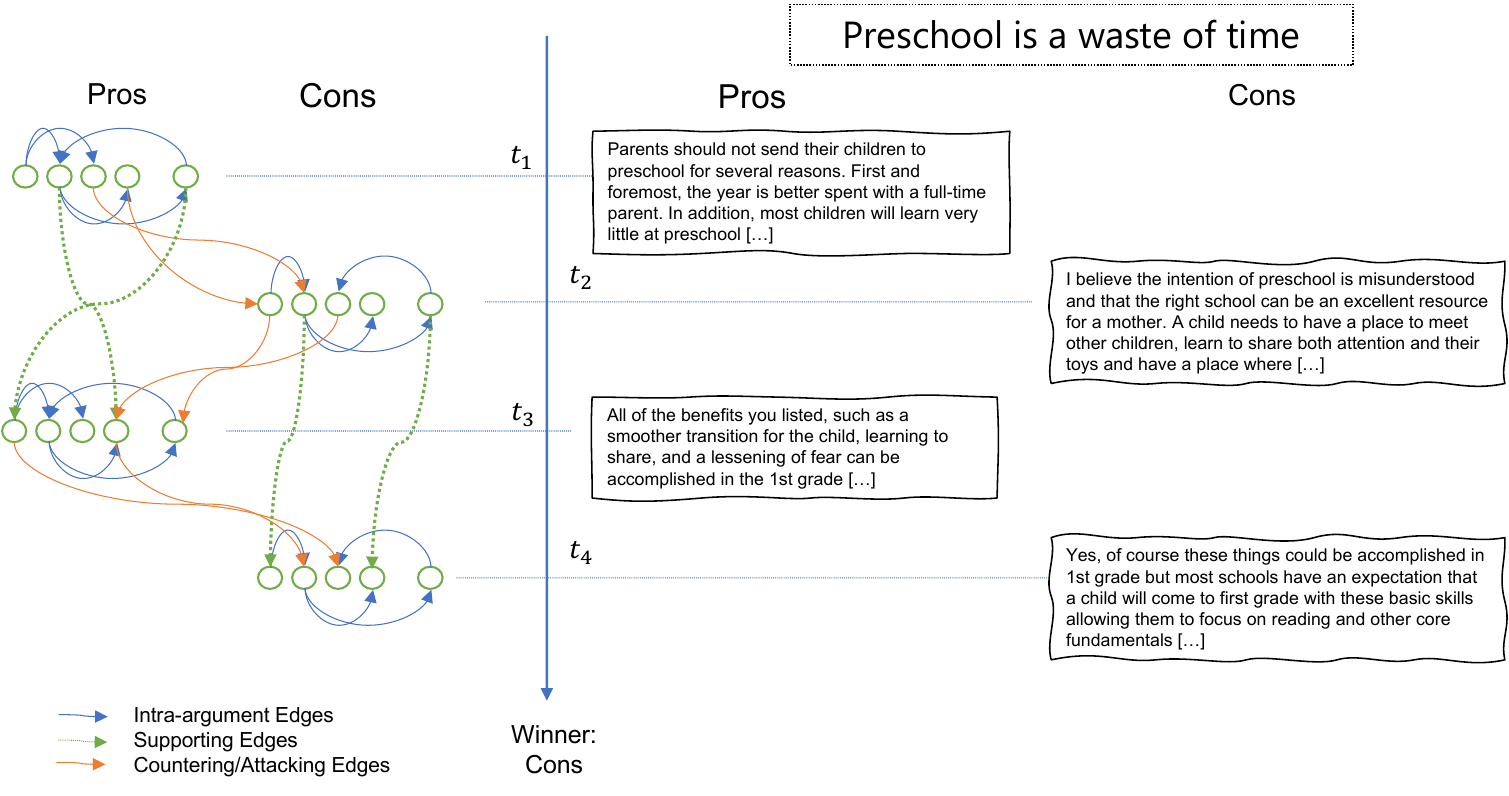}
    \caption{Graph Construction from Debate: Nodes establish connections through three distinct edge types, indicated by colored arrows. Intra-argument edges (blue) link nodes within the same turn, reinforcement edges (green) connect nodes from the same debater across different turns, while countering edges (orange) connect nodes from a debater to their opponent's, illustrating counter-argumentation. The sample debate is taken from data collected by \cite{durmus2019corpus}.}
    \label{fig:graph}
\end{figure*}
Before describing the details of the proposed method, we first give a brief introduction to how we construct a graph for an online debate.
\subsection{Debate Format} Our primary focus lies in online debates wherein the victor emerges through the collective votes of an audience or a panel of judges. These debates adhere to the Oxford-style format, featuring two participants representing opposing viewpoints —one in favor of the claim (Pros) and the other in opposition (Cons) — who alternate in presenting their arguments on a given topic. After the debate, a winner is declared, unless a tie occurs. In this study, we define a \textit{turn} as each instance when a debater presents their argument, and a \textit{round} represents the stage in which opposing sides provide their arguments. Consequently, round 0 consists of turn 0 and turn 1, round 1 consists of turn 2 and turn 3, and so forth.

\subsection{Debate-to-Graph construction}
Given a debate that contains a total of $N$ sentences, a directed, unweighted graph $\mathcal{G}=(\mathcal{V},\mathcal{E}, {\mathcal{H}})$  is constructed based on sentences and their relationships (Figure ~\ref{fig:graph}). Sentences in the debate are represented by a set of nodes $\mathcal{V}$ ($|\mathcal{V}| = N$), and a node attribute matrix $\mathcal{H} \in R^{N\times D}$, defined by $D$-dimensional embedding vectors for each of the sentences. Sentences in the debate may be interconnected and these interconnections are represented by $\mathcal{E}$, the set of edges in the graph.

\textbf{Edge types:} We define three different types of edges to elucidate the participants' strategies throughout the debate. Each type is categorized based on the turn it corresponds to and the strategic role it plays. In Section \ref{sec: method}, we will delve into how each type contributes to node feature aggregation.
\begin{enumerate}
    \item Logical and Coherent Edges: These edges emphasize the participants' ability to construct logical and coherent arguments within their turn.
    \item Reinforcement Edges: These edges serve to strengthen the points previously made by the debater in their previous rounds. We will interchangeably use the terms \textit{reinforcement edges} and \textit{supporting edges}.
    \item Counterargument Edges: These edges highlight the participants' skill in countering their opponents' arguments effectively.
\end{enumerate}

\indent \textbf{Intra-argument Links} These edges connect sentences of the same turn. During a turn, edges are constructed based on the relative position among sentences. These  \textit{Logical and Coherent} edges capture coherency in an argument turn. Given two sentences, denoted as $s^t_i$ and $s^t_j$, both belonging to turn $t$, we establish an edge $e^{inter}_{ij}$ from $s^t_j$ to $s^t_i$ if the positional difference $\mathcal{D}$ between them is within a specified distance threshold $d$. 
$$    e^{intra}_{ij} = 
    \begin{cases}
        1 & \text{if } \mathcal{D}(s^t_i, s^t_j) \leq d\\
        0 & \text{otherwise}
    \end{cases}$$
\indent \textbf{Cross-argument Links} These edges interconnect sentences that belong to different turns and are categorized into two types: \textit{Reinforcement} and \textit{Counterargument} edges. The former connects nodes belonging to the same debater whereas the latter connects nodes belonging to different debaters. For example, nodes in the 3rd turn are connected to nodes from the 1st turn through \textit{Reinforcement} edges and are also linked with their opponent's nodes from the 2nd turn. Unlike intra-argument edges that rely on the relative positions of sentences, cross-argument edges are established using semantic textual similarity between sentences. In this work, we use cosine similarity $S_c$ to capture the semantic relationship of texts. An edge $e_{ij}$ links 2 nodes $v_i$ and $v_j$ if their similarity score $S_c(\mathbf{h}_i, \mathbf{h}_j)$ meets a threshold value $S_{th}$
$$    e^{inter}_{ij} = 
    \begin{cases}
        1 & \text{if } S_c(\mathbf{h}_i, \mathbf{h}_j) \geq S_{th} \\
        0 & \text{otherwise}
    \end{cases}$$
where $\mathbf{h}_i$ and $\mathbf{h}_j$ are $i^{th}$ and $j^{th}$ rows in $\mathcal{H}$, representing embedding vectors of sentences $v_i$ and $v_j$, respectively. $S_{th}$ serves as a crucial hyper-parameter for evaluating the influence of participant interactions on the debate's outcome. An alternative approach is to employ the top $k$ similarities, allowing each node to establish connections with up to $k$ cross-argument nodes that possess the highest similarity scores. We will evaluate the effectiveness of each approach on the predictive performance in Section \ref{sec:evaluation}. It is important to note that cross-argument edges consistently flow from nodes in previous turns to nodes in subsequent turns; there is no reverse direction.

\section{Proposed method} \label{sec: method}
\subsection{Utterance Encoder}
We encode each sentence using pre-trained sentence embedding (Sentence Transformer (SBERT)) \cite{reimers2019sentence}. In preliminary work, we found that this approach works better than using GloVe \cite{pennington2014glove} word embeddings and a bidirectional LSTM to encode semantic vectors for sentences. This step gives us the sentence embedding matrix $\mathcal{H}$, in which each row  $\mathbf{h}_i$ is an embedding vector for sentence $s_i$.

\textbf{Turn Embeddings}: Participants employ distinct strategies during different debate turns. For instance, in the initial round consisting of two turns, the first participant presents their perspective on the topic while the second participant challenges their opponent's arguments and introduces their own viewpoint. We incorporate the temporal turn information into the node features by concatenating it with the sentence embedding $\mathbf{h}_i$. We opt for a 30-dimensional embedding vector $\mathbf{h}_{it} \in R^{30}$ to represent the turn information for each node.
\begin{equation}\label{eq:utterance}
    \mathbf{h}_i = \mathbf{h}_i \Vert \mathbf{h}_{it}
\end{equation}
Let $B$ denote the number of dimensions of the embedding vector of a sentence from SBERT, then $D = B + 30$. 
\subsection{Information flow}
\textbf{Graph Attention Layer}: We employ a Graph Attention Network (GAT) \cite{velivckovic2017graph} layer to update the node representation. The attention mechanism allows GAT to focus on and weigh the importance of different neighbors when aggregating information for each node, called the ``attention score". We are motivated to use GAT in our model because, intuitively, not all sentences in the debate carry equal importance. One can detect the opponent's argumentative ``vulnerable region" \cite{jo2018attentive} and effectively counter it to win the debate. This layer takes as input a set of $A$ ($A \leq N$) node features $\mathbf{h}\in \mathbb{R}^{A \times D}$ and produces a new set of node features $\mathbf{h'}\in \mathbb{R}^{A \times D'}$ ($D' < D$). The attention score of sentence $j$ to sentence $i$ is computed as:
$$    \alpha_{ij} = \frac{\exp(\mbox{LeakyReLU}(\mathbf{a}^T[\mathbf{W}\mathbf{h}_i || \mathbf{W}\mathbf{h}_j]))}{\sum_{k \in \mathcal{N}} \exp(\mbox{LeakyReLU}(\mathbf{a}^T[\mathbf{W}\mathbf{h}_i || \mathbf{W}\mathbf{h}_k])}$$
where $\mathbf{W} \in \mathbb{R}^{D\times D'}$ and $\mathbf{a}\in \mathbb{R}^{2D'}$ are trainable weight matrix and vector of the layer. The output features of node $i$ is the weighted sum of the features of its neighboring node set $\mathcal{N}_i$: 

$$\mathbf{h}'_i = \sum_{j\in \mathcal{N}_i}\alpha_{ij}\mathbf{W}\mathbf{\mathbf{h}_j}$$
In this work, we employ three distinct GAT layers, each responsible for aggregating information from a specific type of edge. We refer to these layers as GATI (intra-argument edge), GATC (counterargument edge), and GATS (supporting edge). At each turn, the GAT layer processes a specific set of input node features and produces a new set of features, called \textit{interaction} representation of each sentence: 
\begin{align}
    \mathbf{h}^t_I &= \text{GATI}(\mathbf{h}_{\mathcal{I}_t}; \mathbf{a}^I, \mathbf{W}^I) \label{eq:gati}\\
    \mathbf{h}^t_C &= \text{GATC}(\mathbf{h}_{\mathcal{J}_t}; \mathbf{a}^C, \mathbf{W}^C) \label{eq:gatc} \\
    \mathbf{h}^t_S &= \text{GATS}(\mathbf{h}_{\mathcal{K}_t}; \mathbf{a}^S, \mathbf{W}^S) \label{eq:gats}
\end{align}
where $\mathbf{a}^{*}$ and $\mathbf{W}^*$ are vectors and matrices associated with each layer. Here, we have three sets of node features: $\mathbf{h}_{\mathcal{I}_t}$, $\mathbf{h}_{\mathcal{J}_t}$, and $\mathbf{h}_{\mathcal{K}_t}$, each corresponding to distinct node sets:
\begin{itemize}
    \item $\mathcal{I}_t$ represents the set of nodes that pertain to the same time step, encompassing nodes within the current turn. $\mathbf{h}_{\mathcal{I}_t} = \{\mathbf{h}^t_1, \mathbf{h}^t_2, \mathbf{h}^t_3, ...\}$ denotes features matrix of a set of nodes at time $t$.
    \item $\mathcal{K}_t$ comprises nodes from time steps $t-2$ and $t$, all originating from the same debater and exhibiting a supportive relationship. This set characterizes argumentative enhancement or promotion. Note that the set of node features at time $t-2$ are \textbf{updated} in turn $t-2$. Therefore, $\mathbf{h}_{\mathcal{J}_t} = \{\mathbf{h}'^{t-2}_1, \mathbf{h}'^{t-2}_2, ..., \mathbf{h}^t_1, \mathbf{h}^t_2, ...\}$ denotes the updated features matrix of a set of nodes at times $t-1$ and utterance matrix of nodes at $t$.
    \item In contrast, $\mathcal{J}_t$ encompasses nodes from time steps $t-1$ and $t$ and signifies an adversarial relation, capturing how a debater challenges an opponent's position by considering nodes from the opponent's previous turn ($t-1$). Because nodes feature at time $t-1$ are updated, $\mathbf{h}_{\mathcal{K}_t} = \{\mathbf{h}'^{t-1}_1, \mathbf{h}'^{t-1}_1, ..., \mathbf{h}^t_1, \mathbf{h}^t_2, ...\}$.
\end{itemize}

\paragraph{Sequential Update}
The node features are updated sequentially using a temporal attention mechanism. Information propagation occurs along \textit{directed} edges, and the features of nodes at time $t$ are updated based on their neighboring nodes from the same turn (via intra-argument edges) as well as nodes from previous turns (via cross-argument edges) (Figure \ref{fig:flow}). This information flow scheme illustrates the cognitive process of a debater during their turn, as they must consider the opponent's previous arguments, formulate counterarguments, reinforce their own points, and even introduce new ideas. The node features updated at time $t$ serve as the input when updating node features at times $t+\tau$ ($\tau \in \{1,2\}$). This process shares similarities with traditional RNNs like LSTM and GRU. However, it is important to note that our work focuses on handling a specific subset of nodes at each timestep. This distinction sets us apart from Gated Graph Sequence Neural Networks \cite{li2015gated} that process the entire graph as input at each timestep. Similar to an RNN-Cell, that operates on a single input element at each time step and generates output that serves as a hidden feature for subsequent times, we introduce the SGA layer to manage the processing of a specific subset of nodes at time $t$. The entire debate graph is processed sequentially subgraph-by-subgraph. 
\begin{figure*}
    \centering
    \includegraphics[scale=0.6]{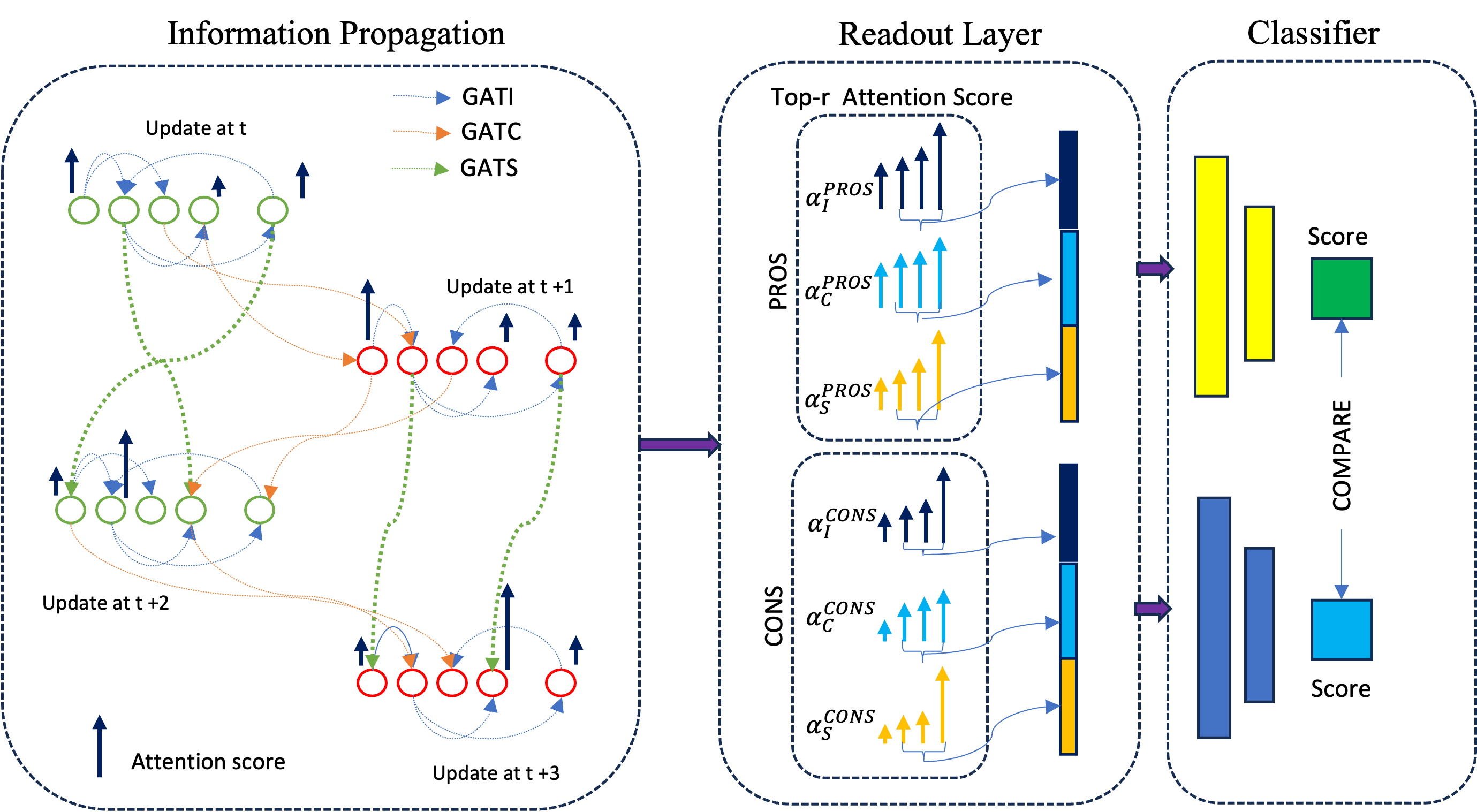}
    \caption{The proposed architecture consists of three key modules: (1) Information propagation is driven by the SGA layers, updating node features sequentially using a graph attention mechanism. (2) The readout layer identifies representative vectors associated with each debater, which are subsequently supplied as input to (3) an MLP classifier for predicting the debate winner.}
    \label{fig:arch}
\end{figure*}

Given a debate $\mathcal{S}$ that has $T$ turns: $\mathbb{S} = \{ S_t; t \in [0,T-1] \}$, $S_t = \{s_j^t; j\in [0,M_t-1]\}$ denotes a debate turn consisting of $M_t$ sentences $s_j^t$. It is noticeable that $N = \sum_{t=0}^{T-1}M_t$. Let $\mathbf{h}_j^t$ the utterance embedding of the sentence $s_j$ (from \ref{eq:utterance}), the new node feature $h'_j$ is calculated using the SGA layer which executes the following operations (we discard the superscript $t$ for readability):
\begin{equation}
    h'_j = SGA(\mathbf{h}_j, \mathbf{h}_{\mathcal{I}}, \mathbf{h}_{\mathcal{J}}, \mathbf{h}_{\mathcal{K}})
    = \mathbf{h}_j \otimes \mathbf{h}^{\text{X }}_j
\end{equation}
where $\otimes$ is the update operator using GRU operations \cite{cho2014learning}. The $\mathbf{h}^{\text{X }}_j$ denotes the interaction representation feature at time $t$, encompassing intra-argument coherency, counterarguments against the opponent's points, and reinfordcement of the debater's previous statements. It is calculated by concatenating the node features produced by three component GAT layers (equations \ref{eq:gati}, \ref{eq:gatc}, \ref{eq:gats}):
\begin{equation}
    \mathbf{h}^{\text{X }}_j = \mathbf{h}^{\text{GATI }}_j || \text{ } \mathbf{h}^{\text{GATC }}_j || \text{ } \mathbf{h}^{\text{GATS }}_j
\end{equation}
It is important to observe that during the initial turn, denoted as $t=0$, there are no counterarguments in the debater's thoughts. As a result, we initialize $\mathbf{h}^{\text{GATC }_0}_j$ to be equal to 0. Additionally, a debater does not introduce a reinforcing argument until their second round (or when $t \geq 2$). Consequently, both $\mathbf{h}^{\text{GATS }_0}_j$ and $\mathbf{h}^{\text{GATS }_1}_j$ are set to 0 during this period. The updated node features $h'_j$ are then employed to update the attributes of nodes in subsequent turns.

\subsection{Readout Layer}
Once all the node features have been updated, we employ a readout layer to ``summarize" the ideas presented by each participant during the debate. For each debater, we select a set of top $r$  (e.g., $r=3$) \textit{representatives}, which are used as input for the prediction classifier. The process of selecting these representative nodes is determined by the highest "attention scores" generated by each GATI, GATC, and GATS layers, denoted as $\alpha_I$, $\alpha_C$, and $\alpha_S$, respectively. During the feature update step, each node receives an attention score from its neighboring nodes. These scores emphasize the significance of a node in relation to others. The more significant a node is, the greater its contribution to a debater's overall idea. The total attention received by each node is obtained by summing up its individual attention scores. Consider a node $s_l$, its attention scores are:

\begin{equation}
    \label{eq:readout}
    \alpha^I_{s_l} = \sum_{i\in \mathcal{I}} \alpha_{i},  
    \text{     }\alpha^C_{s_l} = \sum_{j\in \mathcal{J}} \alpha_{j}, 
    \text{     }\alpha^S_{s_l} = \sum_{k\in \mathcal{K}} \alpha_{k} 
\end{equation}
We opt to select the top $r$ nodes with the highest scores for each type of attention. We then concatenate the feature vectors corresponding to these selected nodes to create a $3 \times r\times D'$-dimensional vector, where $D'$ is the dimension of the node feature produced by SGA. The readout layer subsequently generates two ``summary" vectors, each serving as a deep representation of each debater's performance during the debate. 
\subsection{Classification}
The two vectors, $\mathbf{Q}^{PROS}$ and $\mathbf{Q}^{CONS}$ achieved by the readout layer are fed to the classifier to perform the prediction task. Each vector is mapped to a score value $c \in \mathbb{R}^1$ by linear transformation using a Fully Connected (FC) layer followed by an activation function (e.g., ReLU), Layer Norm (LN) \cite{ba2016layer} and dropout layer \cite{srivastava2014dropout}. Let us denote a series of FC + ReLU + LN + Dropout an MLP, then
\begin{align*}
    c^{PROS} &= \textit{MLP1}(\mathbf{Q}^{PROS}) \\
    c^{CONS} &= \textit{MLP2}(\mathbf{Q}^{CONS})
\end{align*}
If the Pros side wins, we expect that $c^{PROS} > c^{CONS}$, and conversely when the Cons side wins. Here, we denote $C^+$ and $C^-$ as the scores of the winner and loser, respectively. Our objective is to maximize the difference between $C^+$ and $C^-$ as much as possible. To achieve this, we employ Pairwise Cross-Entropy (PCE) loss, that minimizes:
\begin{equation}
    \mathcal{L} = \text{PCE}(C^+, C^-) = \log(1 + exp(C^- -  C^{+}))
\end{equation}
The network architecture is illustrated in Figure \ref{fig:arch}.


\section{Evaluation}\label{sec:evaluation}
\subsection{Dataset}
Our study is conducted on the \textit{debate.org} dataset collected by \cite{durmus2019corpus}. The dataset contains 78,376 debates on controversial topics, including \textit{abortion}, \textit{death penalty}, \textit{gay marriage}, and \textit{affirmative action}. Each debate consists of multiple rounds in which two participants from two opposing sides take turns expressing their opinions. Further details can be found in \cite{durmus2019corpus}.
\paragraph{Winning criterion} The winner is determined by the criterion of ``Made more convincing arguments”. We exclude debates with fewer than 5 voters and tie debates. Additionally, debates in which the winner has just one more vote than the loser are also classified as ties.
\paragraph{Preprocessing} To study the interaction among debates, we only keep debates that have at least 3 rounds (equivalent to 6 turns). Short arguments are also eliminated, i.e., we remove debates that have fewer than 5 sentences in each round (each graph thereby has at least 30 vertices). The first 3 rounds of longer debates are used for analysis. The dataset exhibits an imbalance, with the Cons side accounting for 65\% of the winners whereas the Pros side wins only 35\%. To create a balanced dataset, we also use the final 3 rounds of the debates where the Pros side wins and the debate comprises more than three rounds. This data augmentation step also increases the size of the dataset. 
\paragraph{Statistics} After the experimental dataset selection step, there are a total of 2,445 debates available for model training and testing. Among these debates, the Pros side wins in 1,130 debates, while the Cons side secures victory in 1,325 debates. Additional statistical information is shown in table \ref{tab:stats}. Observing the table, it becomes evident that the winning side tends to produce more sentences and more counterarguments compared to the losing side. Conversely, the losing side appears to prioritize reinforcing their own ideas rather than generating a higher number of counterarguments. 
\begin{table}[]
\centering
\caption{The number of sentences, number of Counterargument Edges, and number of Supporting Edges made by winner and loser in an argument turn. Cross-argument edges are constructed using a similarity threshold of $0.85$.}
\begin{tabular}{c|ccc}
       & \#Sentences & \#Countering        & \#Supporting \\ \hline
Winner & 38.6        & 6.96                      & 5.93               \\
Loser  & 36.1        & 6.78                      & 6.64              
\end{tabular}
\label{tab:stats}
\end{table}
\subsection{Experimental setup} 
\paragraph{Data Preprocessing} We randomly split the dataset with 60\% for training, 20\% for validation and 20\% for testing. For text normalization, we employ the following steps: (1) replacing URLs with ``website", (2) replacing all the numbers with ``number", and (3) lowercasing text. Next, we employed spaCy \cite{spacy} for sentence tokenization. Sentences are then encoded by SBERT's ``all-MiniLM-L6-v2" model that transforms a sentence into a 384-dimensional vector.
\paragraph{Parameter setting} We use a similarity threshold of 0.85 for cross-argument edge construction, other approaches regarding edge construction will be further discussed in the ablation study. The intra-argument distance threshold is $d = 3$. Each node within a turn links to nodes that share a relative positive correlation within a 3-node proximity.  Node features updated by each GAT layer have $D' = 32$ dimension.  For the readout layer, we choose $r = 3$. We use a stack of three MLPs to transform the readout layer's output into a score for each debater. The first layer reduces the vector from $3\times r\times D$ to half its size. The second layer further reduces the output of the first layer by half, and the final layer maps the second output vector to a real value. We apply the tanh function to ensure the value falls within the range [-1; 1]. For hyper-parameters, we apply the dropout rate of 0.2 for all GAT layers and the classifier. Optimization is performed using Adam \cite{kingma2014adam}. The batch size is 32. We run the model for 50 epochs with early stopping. The learning rate is 0.0001.  
\paragraph{Other settings} Deep learning frameworks are Pytorch \cite{pytorch} and Pytorch Lightning \cite{falcon2019pytorch}. We use DGL package \cite{dgl} as the graph deep learning framework. The networks are trained and tested on an NVIDIA Quadro RTX 8000 GPU with 50GB of memory.
\subsection{Comparison baselines} Given that the Cons side accounts for 52.5\% of wins in the test set, it serves as the \textbf{majority baseline}, representing the best prediction one can make regardless of the input features. We compare our model's performance to SOTAs in debate winning prediction which adopt sequence approach in their work. 
\paragraph{Sequence approach} In the study by \cite{hidey2018persuasive}, they aggregate the entire discussion into a single sequence and model it using LSTM with an attention mechanism applied to the sentences, referred to as the \textbf{all-LSTM} approach. They also incorporate implicit discourse relations using the Penn Discourse Tree Bank \cite{prasad2008penn} discourse structure. While their research primarily centers on the Reddit dataset \cite{tan2016winning}, we apply the same methodology to our debate dataset. Additionally, we find relevance in the work of \cite{li2020exploring}, denoted as \textbf{ASODP}, which shares our focus on Oxford-style debates and employs a sequential approach for debate analysis. Furthermore, \cite{zeng2020changed} introduces the \textbf{DTDMN} method, designed to process pairs of conversations and predict their persuasiveness. Similarly, we present the Pros and Cons sides as inputs to facilitate comparative analysis.
\paragraph{Graph approach} To highlight the significance of processing the debate on a turn-by-turn basis, we introduce two baseline models for graph analysis. The first baseline employs a 2-layer \textbf{GAT} network, while the second baseline utilizes a \textbf{GGNN}. These GNNs serve as information aggregators and feature extractors for the debate graph, simultaneously processing all nodes in the graph (and repeating this process 6 times, corresponding to 6 turns in the case of GGNN). In the case of GAT, the initial layer transforms the input into 64-dimensional vectors, and the subsequent layer maps the output from the first layer to 32-dimensional features. In GGNN, we also utilize a 32-dimensional output feature size to align with the output feature size of our SGA layer. To summarize the node features for each debater, we introduce a mean readout operation.
\paragraph{Temporal graph approach} Since no other sequential graph approach exists for debate winning prediction, we adopt the information flow method proposed in \cite{chen2019graphflow} (\textbf{Graphflow}), initially designed for machine comprehension. We utilize the output of the RGNN layer from the final turn, feeding it into the MLP layer for the prediction task.
\subsection{Experimental results}
\begin{table}[]
\caption{Debate-winning prediction results. The best results are in bold. (**: Using the top 3 highest similarity scores to construct cross-argument edges, *: Using a threshold value of 0.85 to construct cross-argument edges).}
\centering
\begin{tabular}{|l|ll|}
\hline
\multicolumn{1}{|c|}{\textbf{Models}}                  & \multicolumn{1}{c}{\textbf{Acc.}} & \textbf{F1}    \\ \hline
\multicolumn{1}{|l|}{{{\textbf{Majority Baseline}}}} & \multicolumn{1}{c}{0.525}              &                \\ \hline
\multicolumn{1}{|l|}{{\textbf{Sequence Baseline}}} & \multicolumn{1}{c}{}              &                \\
all-LSTM                                            & 0.635                             & 0.563          \\
ASODP                                                  & 0.656                             & 0.623          \\
DTDMN                                                  & 0.660                             & 0.625          \\ \hline
{{\textbf{Graph Baseline}}}                          &                                   &                \\
GAT                                                   & 0.541                              & 0.472           \\
GGNN                                                    & 0.565                              & 0.522           \\ \hline
{{\textbf{Sequence Graph Baseline}}}     &    &  \\
Graphflow & 0.645  & 0.620 \\ \hline
{{\textbf{SGA}}}                                     &                                   &                \\
w/o GATI                                               & 0.621                             & 0.523          \\
w/o GATC                                               & 0.562                             & 0.495          \\
w/o GATS                                               & 0.629                             & 0.534          \\
FULL MODEL                                             &                                   &                \\
$^{*}$S = 0.85                                              & 0.654                             & \textbf{0.667}          \\
$^{**}$k = 3                                                &  \textbf{0.675}                            & 0.625 \\ \hline
\end{tabular}
\label{tab:results}
\end{table}
\begin{figure}
    \centering
    \includegraphics[scale=0.35]{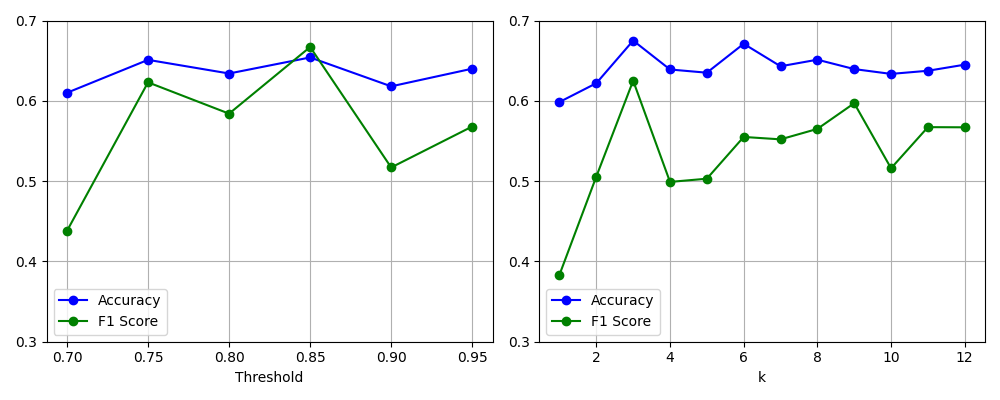}
    \caption{Impact of cross-argument construction values on network performance. Left: Edge construction using a threshold value. Right: Edge construction using top-k highest values.}
    \label{fig:thres}
\end{figure}
The evaluation results are presented in  Table \ref{tab:results}.  The sequence baselines (all-LSTM, ASODP, DTDMN) all perform similarly, with DTDMN producing the best accuracy of this group at 66.0\%. The Graph baselines perform more poorly, with the highest accuracy, 56.5\% produced by SSGN. Graphflow, boasting an accuracy of 64.5\%, outperforms traditional graph approaches. However, it still trails behind the robust benchmarks set by sequential approaches such as ASODP and DTDMN. Our full model, SGA with k=3, outperforms all baselines with an accuracy of 67.5\%, a 1.5\% absolute (2.3\% relative) improvement over DTDMN.  The F1-score, achieved by constructing cross-argument edges with a threshold of 0.85, significantly outperforms the baselines. It reaches 66.7\%, representing a 4.2\% absolute (or 6.7\% relative) improvement over DTDMN. We thus demonstrate that we outperform state of the art models for this dataset. 

The performances of all-LSTM and DTDMN are diminished when applied to the \textit{debate.org} dataset. This can be attributed to a fundamental distinction between the two domains. In the context of \textit{debate.org}, the ultimate determination of the winner is not based on subjective criteria but rather relies on the judgments of a panel of judges or the voters. The voters place substantial emphasis on the debaters' ability to rigorously address and counter their opponents' reasoning. Furthermore, they favor debaters who engage in high-quality and dynamic interactions throughout the debates.
\paragraph{Sequence matters} The results show a significant superiority of sequence-based baselines over graph-based ones when applied to the debate dataset. This highlights the critical significance of adopting a sequential approach, where the debate is processed turn-by-turn, rather than relying solely on graph-based methodologies.
\paragraph{Counter-argument is crucial} We extended our analysis by performing an ablation study to assess the individual impact of each GAT layer on our proposed SGA model. We observe that when we omit the counter-argument edges, the reduction in network performance was more significant compared to scenarios where we exclude either GATI or GATS layers. Specifically, accuracy drops by 11.3\%, in contrast to 5.4\% and 4.6\%, respectively. This outcome can be elucidated by considering that if a debater disregards the opponent's remarks from the preceding turn, their persuasive ability may diminish in the eyes of the voters or judges. In essence, acknowledging and responding to counter-arguments plays a pivotal role in constructing compelling arguments in a debate context.
\subsection{Impact of graph parameters}
We conduct a detailed analysis of the impact of graph construction parameters, such as $S_{th}$ and $k$, on the network's performance (Figure \ref{fig:thres}). In the context of employing a similarity threshold, it is noteworthy that a threshold value of $0.85$ yields the highest performance in terms of accuracy and F1-score. 

Regarding the top-k approach, it is worth highlighting that while $k=3$ achieves the highest accuracy, as well as highest F1-score. These insights into parameter effects contribute to a deeper understanding of how to optimize network performance for specific objectives and trade-offs.
\section{Related work} \label{sec: related}
Graph Neural Networks (GNNs) have proven to be powerful tools for harnessing insights into, and making predictions on, data structured as graphs, particularly in the realm of Natural Language Processing (NLP). Within NLP, GNNs have been applied to a wide spectrum of tasks including, but not limited to, dependency parsing \cite{ji2019graph}, sentiment analysis \cite{liang2022aspect}\cite{li2021dual}, and semantic understanding \cite{kipf2017semi}. In recent developments, researchers in NLP have extended GNNs by integrating them with RNNs to enable sequential processing of graph-structured data. Notably, \cite{song2018graph} introduced a graph-to-sequence methodology for the AMR-to-text generation task, wherein they construct an Abstract Meaning Representation (AMR) graph and progressively update the entire graph during sequential generation. Furthermore, \cite{chen2019graphflow} made significant strides in the domain of machine comprehension by incorporating conversation history into their model. They adopt a graph-based approach, constructing a graph that evolves with each conversational turn. While our work shares a commonality in the sequential update of subgraphs, it is important to emphasize that the implementation details diverge significantly. Researchers have explored temporal graph approaches for tasks like traffic flow forecasting ~\cite{yu2017spatio}\cite{guo2019attention} and skeleton-based action recognition~\cite{shi2019two}. However, the utilization of sequence graph approaches in conversation analysis, particularly within online debate and argumentative analysis contexts, remains relatively unexplored.

\section{Conclusion and future work} \label{sec: conclusion}
In conclusion, the task of modeling online debates, characterized by the dynamic exchange of ideas, is a challenging endeavor. To tackle this complexity, we introduced a novel approach using sequence-graph modeling. By representing conversations as graphs, we effectively captured the interactions among participants through directed edges, while the sequential propagation of information along these edges enriched our understanding of context. Our incorporation of the SGA layer demonstrated the efficacy of our information update scheme. Our experimental results demonstrate the success of sequence graph networks in outperforming existing methods when applied to Oxford-style online debate dataset.

The proposed method not only advances the ability to model dynamic discussions but also highlights the potential of sequence-graph approaches for a wide range of tasks involving sequential interactions and context-rich data. As online debates continue to evolve, the techniques presented in this paper offer valuable insights into improving our understanding of complex conversational dynamics.

While the proposed method has demonstrated promising results in predicting debate outcomes, it does exhibit certain limitations. Firstly, the construction of cross-argument edges relies solely on similarity scores. While this approach may suffice for reinforcing connections, it may not consistently identify valid counterarguments. High similarity scores between two sentences do not guarantee a counterrelation. Secondly, the method overlooks the utilization of argument structures. The intra-argument links primarily capture temporal relationships by connecting adjacent sentences. However, this approach fails to account for potential relationships between sentences that are distant within an argument turn. There is room for improvement by incorporating pre-trained models that account for argumentative structures. For instance, \cite{li2020exploring} enhanced predictability on debate datasets by integrating argument structure introduced by \cite{niculae2017argument}. 

\section*{Acknowledgment}
This material is based upon work supported by the National Science Foundation under Award \# OIA-1946391, ``Data Analytics that are Robust and Trusted (DART)".

\end{document}